\documentclass{article}
\usepackage{ijcai23}

\usepackage{times}
\usepackage{soul}
\usepackage{url}
\usepackage[hidelinks]{hyperref}
\usepackage[utf8]{inputenc}
\usepackage[small]{caption}
\usepackage{graphicx}
\usepackage{amsmath}
\usepackage{amsthm}
\usepackage{booktabs}
\usepackage{algorithm}
\usepackage{algorithmic}
\usepackage{subfigure}
\usepackage{bm}
\usepackage{multirow}
\usepackage{mathtools}
\usepackage{amssymb}
\usepackage[switch]{lineno}

\title{Pyramid Diffusion Models For Low-light Image Enhancement}

\author{
    Dewei Zhou \and Zongxin Yang \and Yi Yang\footnote{Yi Yang is the corresponding author.}\\
    \affiliations
    ReLER, CCAI, Zhejiang University
    \emails
    \{zdw1999, yangzongxin, yangyics\}@zju.edu.cn
}

\def\eg{\emph{e.g.}}
\def\ie{\emph{i.e.}}

\begin{document}

\maketitle

\begin{abstract}
    Recovering noise-covered details from low-light images is challenging, and the results given by previous methods leave room for improvement. Recent diffusion models show realistic and detailed image generation through a sequence of denoising refinements and motivate us to introduce them to low-light image enhancement for recovering realistic details. However, we found two problems when doing this, \ie, \textbf{1)} diffusion models keep constant resolution in one reverse process, which limits the speed; \textbf{2)} diffusion models sometimes result in global degradation (\eg, RGB shift). To address the above problems, this paper proposes a \textbf{Py}ramid \textbf{Diff}usion model (PyDiff) for low-light image enhancement. PyDiff uses a novel pyramid diffusion method to perform sampling in a pyramid resolution style (\ie, progressively increasing resolution in one reverse process). Pyramid diffusion makes PyDiff much faster than vanilla diffusion models and introduces no performance degradation. Furthermore, PyDiff uses a global corrector to alleviate the global degradation that may occur in the reverse process, significantly improving the performance and making the training of diffusion models easier with little additional computational consumption. Extensive experiments on popular benchmarks show that PyDiff achieves superior performance and efficiency. Moreover, PyDiff can generalize well to unseen noise and illumination distributions. Code and supplementary materials are available at \url{https://github.com/limuloo/PyDIff.git}.

\end{abstract}

\section{Introduction}

Low-light images suffer from noise bursts, and recovering ideal normal-light images from them is a long-studied problem. Thanks to the development of deep learning, many effective methods have been proposed. LLNet~\cite{lore2017llnet} and SID~\cite{chen2018learning} show the power of neural networks by training them on lots of paired data. According to the Retinex theory~\cite{land1977retinex}, KIND~\cite{zhang2019kindling} and RetinexNet~\cite{wei2018deep} decompose the illumination and reflectance map through a well-designed loss. To handle this highly ill-posed problem, LLFLOW~\cite{wang2022low} introduces normalizing flow models~\cite{kingma2018glow} to low-light image enhancement.

\begin{figure}[tb]
    \centering
        \includegraphics[width=0.45\textwidth]{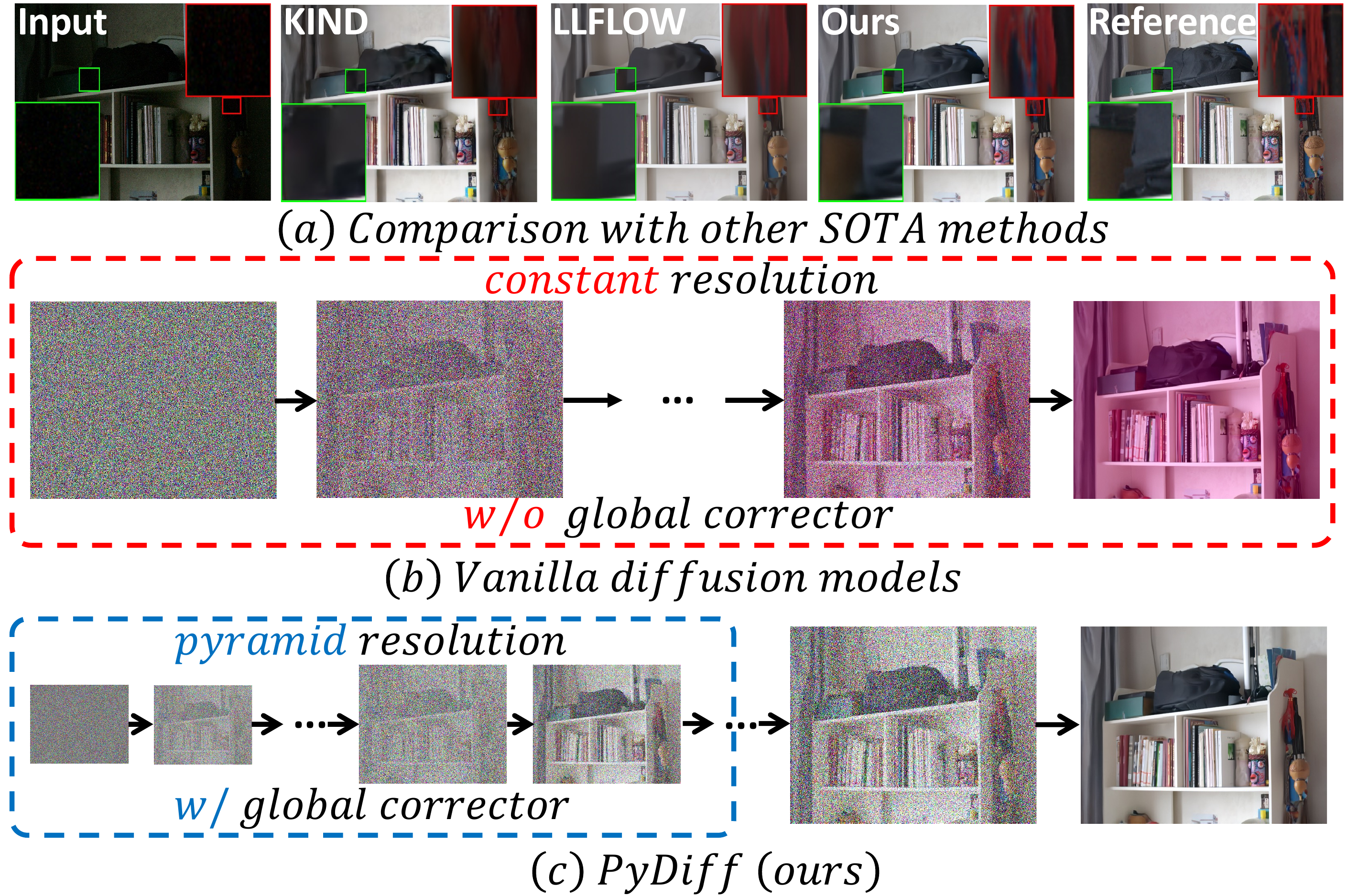}
    \caption{(a) Compared with other SOTA methods, our PyDiff generates more realistic details and restores correct colors. For better viewing, we brighten the Input. (b) Vanilla diffusion models perform sampling in a constant resolution style, and they result in global degradation similar to the RGB shift we analyze in Fig.~\ref{fig:whyHue}. (c) Our PyDiff performs sampling in a pyramid resolution style (\ie, progressively increasing resolution in \textbf{one} reverse process) to achieve faster speed (\ie, to sample at a lower resolution is faster). With the help of a global corrector, PyDiff shows stunning results without global degradation. \textbf{Please zoom in for the best view.}
} 
   
    \label{fig:PyDiff_intro}
  \end{figure}

Although the above methods have made significant progress in low-light image enhancement, noise-covered details restored by them can be further enhanced. As shown in Fig.~\ref{fig:PyDiff_intro}(a), previous methods often lead to blurred details and distorted colors. Diffusion models~\cite{ho2020denoising,song2020score} have recently shown their talents in image generation, which can generate more realistic details through a sequence of refinements. Therefore, we introduce diffusion models to low-light image enhancement for better restoring noise-covered details, as shown in Fig.~\ref{fig:PyDiff_intro}(a).

When introducing diffusion models to low-light image enhancement, we found two problems, as demonstrated in Fig.~\ref{fig:PyDiff_intro}(b). \textbf{1)} The resolution is constant in \textbf{one} reverse process, which limits the speed. \textbf{2)} Diffusion models result in global degradation similar to the RGB shift we analyze in Fig.~\ref{fig:whyHue}.

To solve these problems, we propose a \textbf{Py}ramid \textbf{Diff}usion model (PyDiff) for low-light image enhancement. As shown in Fig.~\ref{fig:PyDiff_intro}(c), PyDiff uses a novel pyramid diffusion method to sample images in an efficient pyramid resolution style (\ie, progressively increasing resolution in \textbf{one} reverse process). 
Performing noisier sampling at lower resolution makes the reverse process faster and provides PyDiff with a larger inception field, which benefits global information recovery. Moreover, we analyze the cause of global degradation (Fig.~\ref{fig:whyHue}) and argue that denoising networks are hard to treat global degradation as part of the noise and correct it during denoising since the reverse process is biased to eliminate Gaussian noise.
To alleviate the global degradation that denoising networks can not notice, PyDiff performs sampling with a global corrector. With little additional computational consumption, the global corrector significantly improves the performance and makes the training of diffusion models easier.

We conduct extensive experiments on two popular benchmarks (\ie, LOL~\cite{wei2018deep} and LOLV2~\cite{yang2021sparse}) to validate the effectiveness and efficiency of PyDiff. Experimental results show that PyDiff achieves superior performance quantitatively and qualitatively under various scenarios. Compared to the previous state-of-the-art (SOTA) method LLFLOW, which also requires iterative refinements, PyDiff significantly outperforms LLFLOW with a speed of nearly $2\times$ faster. In particular, when dealing with unseen noise distributions, PyDiff significantly outperforms other SOTA competitors, \eg, \textbf{10} points (SSIM) higher than the second place (NE~\cite{jin2022unsupervised}). When handling unseen illumination distributions, PyDiff also gives competitive results, demonstrating our generalization ability further. 

Our contributions can be summarized below:

\begin{itemize}
    \item To the best of our knowledge, we are the first to introduce diffusion models to low-light image enhancement and achieve SOTA. Using a novel pyramid diffusion method, PyDiff is nearly twice as fast as the previous SOTA method LLFLOW.
    \item We propose a global corrector to alleviate the global degradation that occurs in the reverse process. This significantly improves the performance and makes the training of diffusion models easier with little additional computational consumption.
    \item Experiments on popular benchmarks show that PyDiff achieves new SOTA performance, and PyDiff can generalize well to unseen noise and illumination distributions.
\end{itemize}

\section{Related Work}

\subsection{Low-light image enhancement}

Low-light image enhancement has been studied for a long time, with numerous deep learning-based approaches proposed. LLNet~\cite{lore2017llnet} and SID~\cite{chen2018learning} collect lots of low/normal-light image pairs to train the network. For getting illumination and reflectance maps~\cite{land1977retinex}, RetinexNet~\cite{wei2018deep}, KIND~\cite{zhang2019kindling}, and KIND++~\cite{zhang2021beyond} carefully design the loss to train a decomposition network. EnlightenGAN~\cite{jiang2021enlightengan}, ZeroDCE~\cite{guo2020zero}, and NE~\cite{jin2022unsupervised} propose effective unsupervised methods which do not require paired data. BREAD~\cite{guo2022low} decouples the entanglement of noise and color distortion. Some works~\cite{fan2022half,cui2022illumination,kim2021representative} have brought performance improvements by designing novel and efficient networks. LLFLOW~\cite{wang2022low} models this ill-posed problem via a normalizing flow model~\cite{dinh2016density,kingma2018glow}. Although the above methods have made significant progress in low-light image enhancement, noise-covered details restored by them can be further enhanced. This paper introduces diffusion models~\cite{he2020conditional} to low-light image enhancement for better recovering the details.

\begin{figure}[tb]
    \begin{center}
        \includegraphics[width=0.45\textwidth]{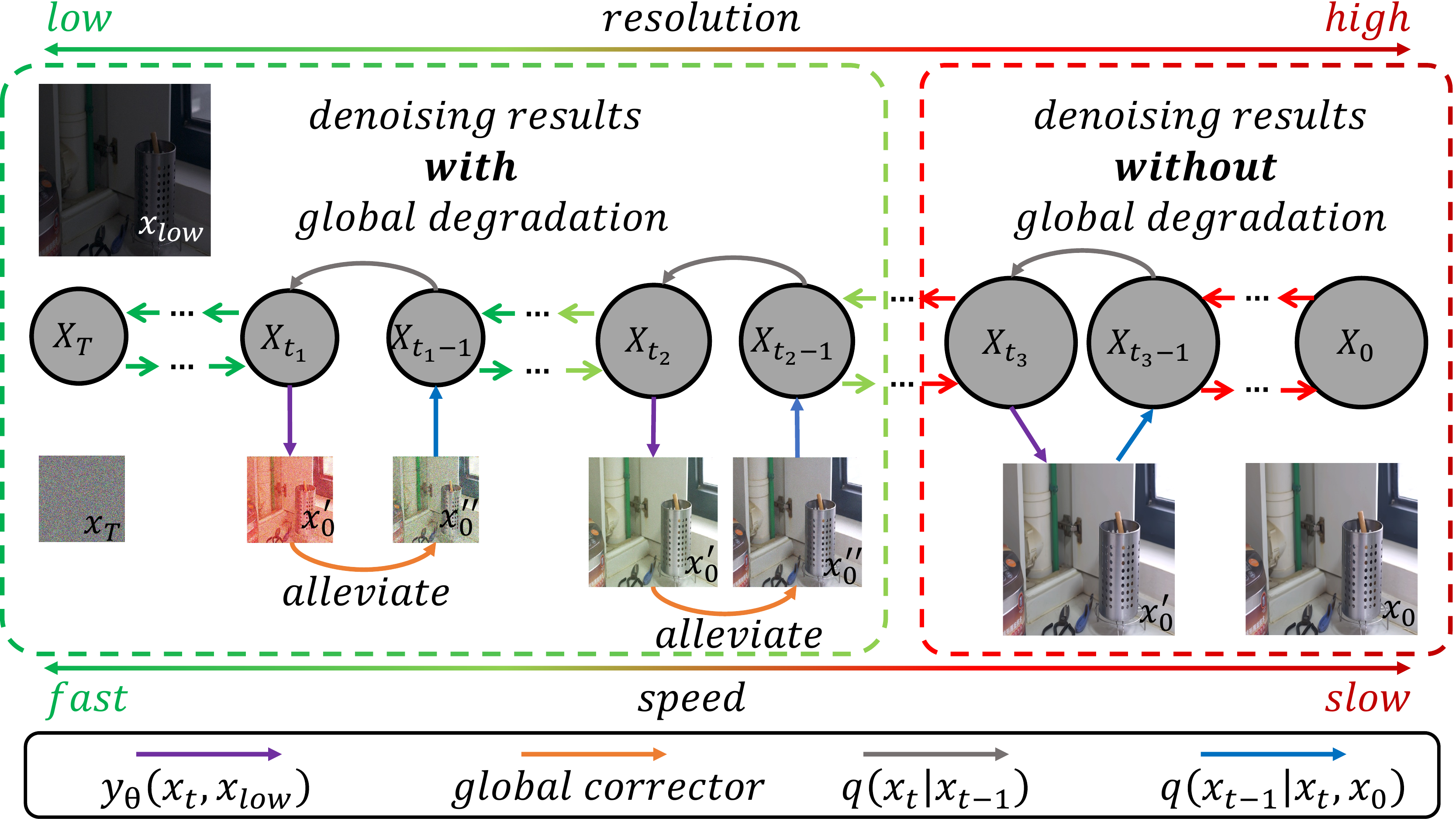}
    \end{center}
    
   \caption{Overview of proposed PyDiff. $y_{\theta}(\mathbf{x}_{t}, \mathbf{x}_{low})$ is the approximate value of $\mathbf{x}_{0}$ calculated according to the denoising network, as discussed in Eq.~\eqref{houyan_theta_cond}. For better viewing, we brighten the $\mathbf{x}_{low}$. \textbf{Please zoom in for the best view.}} 
    \label{fig:overview}
  \end{figure}

\subsection{Diffusion models}
Diffusion Models~\cite{ho2020denoising,song2020score} present high-quality image synthesis results recently. 
However, they typically require a high number of iterations, resulting in slow performance. A number of training-free samplers~\cite{song2020denoising,nichol2021improved,bao2022analytic,lu2022dpm} have been proposed that can achieve comparable results with fewer denoising iterations. 
To achieve conditional generation, Guided-Diffusion~\cite{dhariwal2021diffusion} samples with classifier guidance, while our PyDiff concatenates noisy images with source images to guide denoising like some low-level vision methods~\cite{saharia2022image,saharia2022palette,whang2022deblurring}.

To generate high-resolution images more efficiently, some works~\cite{saharia2022image,ho2022cascaded,fan2022frido} use multiple diffusion models to achieve cascaded high-resolution image synthesis, while LDM~\cite{rombach2022high} makes reverse processes situated within the image encoder's latent space. In \textbf{one} reverse process, the above methods perform sampling at a constant resolution style, limiting the speed. In this paper, PyDiff uses a pyramid diffusion method to achieve faster speed and a global corrector to ensure the sample quality for low-light image enhancement.

\section{Background: Denoising Diffusion Probabilistic Models}

\label{background}
The Denoising Diffusion Probabilistic Model~\cite{ho2020denoising,song2020denoising} is a latent variable model specified by a T-step Markov chain. It starts with a given data distribution $\mathbf{x}_{0} \sim q(\mathbf{x}_{0})$ and repeatedly adds Gaussian noise according to $q(\mathbf{x}_t | \mathbf{x}_{t-1})$ as follows:
\begin{equation}
    q\left(\mathbf{x}_t | \mathbf{x}_{t-1}\right) \coloneqq \mathcal{N}\left(\mathbf{x}_t ; \sqrt{\alpha_t} \mathbf{x}_{t-1},\left(1-\alpha_t\right) \mathbf{I}\right),
\label{eq:ddpm_forward_ori}
\end{equation}
where $\alpha_{t} \in (0, 1)$, and $\alpha_{t} \ge \alpha_{t+1}$.
Using the notation $\bar{\alpha}_{t} \coloneqq \prod_{i=1}^t \alpha_{i}$, the marginal distribution $q(x_{t} | x_{0})$ can be expressed as follows:
\begin{equation}
q\left(\mathbf{x}_t | \mathbf{x}_{0}\right) \coloneqq \mathcal{N}\left(\mathbf{x}_t ; \sqrt{\bar{\alpha}_t} \mathbf{x}_{0},\left(1-\bar{\alpha}_t\right) \mathbf{I}\right)
\label{eq:ddpm_forward}
\end{equation}
When $\sqrt{\Bar{\alpha}_{T}}$ is close to 0, the defined forward process will transform this data distribution into an isotropic Gaussian distribution.

In practical applications, the reverse process of diffusion models is used more often, which converts an isotropic Gaussian distribution to a target data distribution. It is worth mentioning that $q(\mathbf{x}_{t-1} | \mathbf{x}_{t})$ is hard to estimate while $q(\mathbf{x}_{t-1} | \mathbf{x}_{t}, \mathbf{x}_{0})$ is tractable. We can derive the posterior distribution of $\mathbf{x}_{t-1}$ given $(\mathbf{x}_{t},\mathbf{x}_{0})$ with some algebraic manipulation:
\begin{linenomath}
\begin{align}
    q\left(\mathbf{x}_{t-1} | \mathbf{x}_t, \mathbf{x}_0\right) &\coloneqq \mathcal{N}\left(\mathbf{x}_{t-1} ; \tilde{\boldsymbol{\mu}}_t\left(\mathbf{x}_t, \mathbf{x}_0\right), \tilde{\beta}_t \mathbf{I}\right),
    \label{houyan}
    \\
    \tilde{\boldsymbol{\mu}}_t\left(\mathbf{x}_t, \mathbf{x}_0\right) 
&\coloneqq \frac{\sqrt{\bar{\alpha}_{t-1}} \beta_t}{1-\bar{\alpha}_t} \mathbf{x}_0+\frac{\sqrt{\alpha_t}\left(1-\bar{\alpha}_{t-1}\right)}{1-\bar{\alpha}_t} \mathbf{x}_t,
    \\
    \tilde{{\beta}}_t &\coloneqq \frac{1-\bar{\alpha}_{t-1}}{1-\bar{\alpha}_t} \beta_{t},
\end{align}
\end{linenomath}
where $\beta_{t}:=1-\alpha_{t}$. 
We have no $\mathbf{x}_{0}$ during testing, but we can calculate its approximate value according to Eq.~\eqref{eq:ddpm_forward}:
\begin{equation}
    \mathbf{y}_{\theta}(\mathbf{x}_{t}):=\frac{1}{\sqrt{\bar{\alpha}_{t}}}(\mathbf{x}_{t}-\sqrt{1-\bar{\alpha}_{t}}\boldsymbol{\epsilon}_{\theta}(\mathbf{x}_{t})),
\label{eq:cal_x0}
\end{equation}
where $\boldsymbol{\epsilon}_{\theta}(\mathbf{x}_{t})$ is the predicted noise derived from the denoising network, and $\mathbf{y}_{\theta}(\mathbf{x}_{t})$  is an approximation of $\mathbf{x}_{0}$ calculated according to $\boldsymbol{\epsilon}_{\theta}(\mathbf{x}_{t})$. Furthermore, we update Eq.~\eqref{houyan} as follows:
\begin{equation}
    p_{\theta}\left(\mathbf{x}_{t-1} | \mathbf{x}_t\right):=\mathcal{N}\left(\mathbf{x}_{t-1} ; \tilde{\boldsymbol{\mu}}_t\left(\mathbf{x}_t, \mathbf{y}_{\theta}(\mathbf{x}_{t})\right), \tilde{\beta}_t \mathbf{I}\right)
\label{houyan_theta}
\end{equation}

\begin{figure}[tb]
    \begin{center}
        \includegraphics[width=0.45\textwidth]{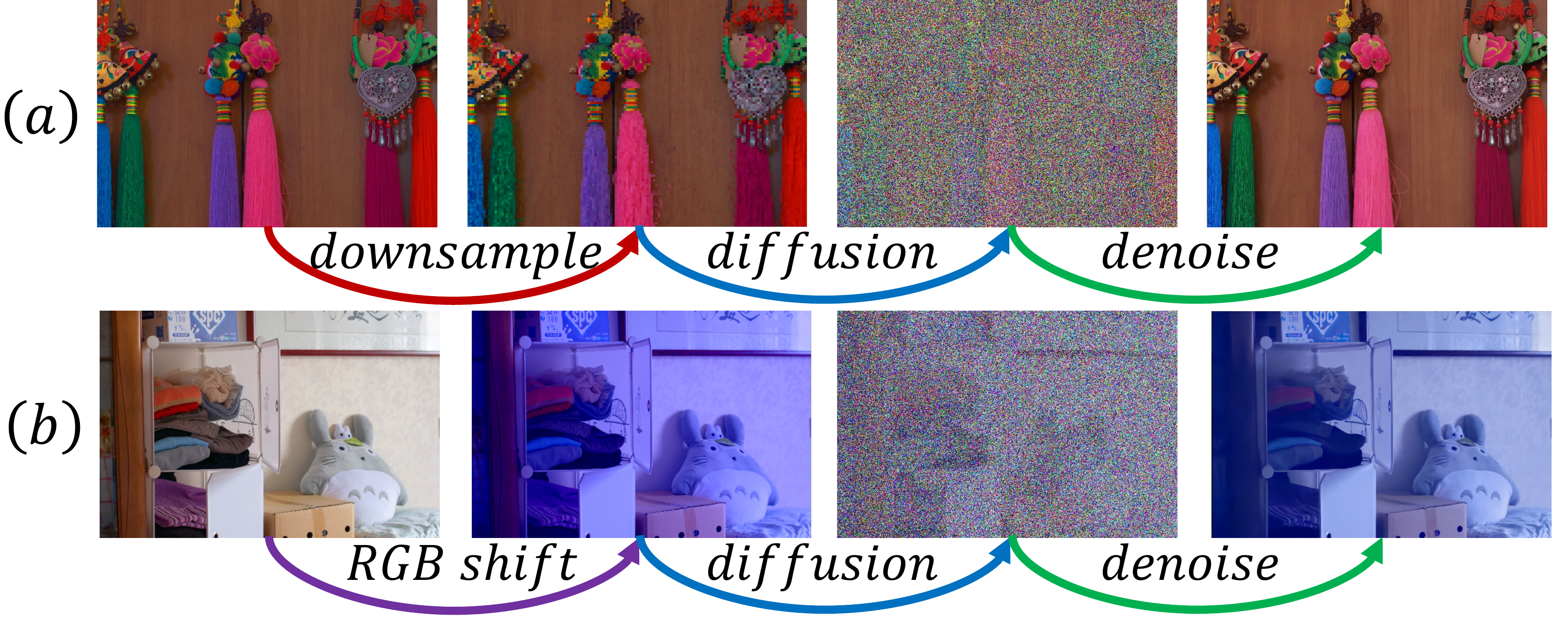}
    \end{center}
    
   \caption{We impose various degradations (\eg, downsampling or RGB shift) on normal-light images and get noisy $\mathbf{x}_{T/2}$ according to Eq.~\eqref{eq:ddpm_forward}. Correspondingly, we begin the reverse process of diffusion from $t=T/2$, conditional on low-light images. We want to know how these degradations affect the second half of the reverse process. (a) Downsampling does not affect the details of the final result. (b) RGB shift will not be corrected. \textbf{Please zoom in for the best view.}} 
    \label{fig:moti}
  \end{figure}

When it comes to image-to-image translation, ~\cite{saharia2022palette,saharia2022image,choi2021ilvr} make the reverse process conditional on an input signal. Specifically, when we need to translate $\mathbf{z}$ to $\mathbf{y}$ (\eg, low-light image to normal-light image), we update Eq.~\eqref{eq:cal_x0} as follow:
\begin{equation}
    \mathbf{y}_{\theta}(\mathbf{x}_{t}, \mathbf{z}):=\frac{1}{\sqrt{\bar{\alpha_{}}_{t}}}(\mathbf{x}_{t}-\sqrt{1-\bar{\alpha}_{t}}\boldsymbol{\epsilon}_{\theta}(\mathbf{x}_{t}, \mathbf{z}))
\label{houyan_theta_cond}
\end{equation}

\nocite{saharia2022image}
\nocite{saharia2022palette}

\section{Methods}

This section presents PyDiff, an effective and efficient method for low-light image enhancement. First of all, we describe the motivation for designing PyDiff. Secondly, we introduce our proposed pyramid diffusion, which significantly improves the inference speed without any performance degradation. Furthermore, we present our proposed global corrector, which alleviates the global degradation that may occur in the reverse process of the diffusion models. Finally, we describe the training and sampling procedures of PyDiff.

\begin{figure}[tb]
    \begin{center}
        \includegraphics[width=0.45\textwidth]{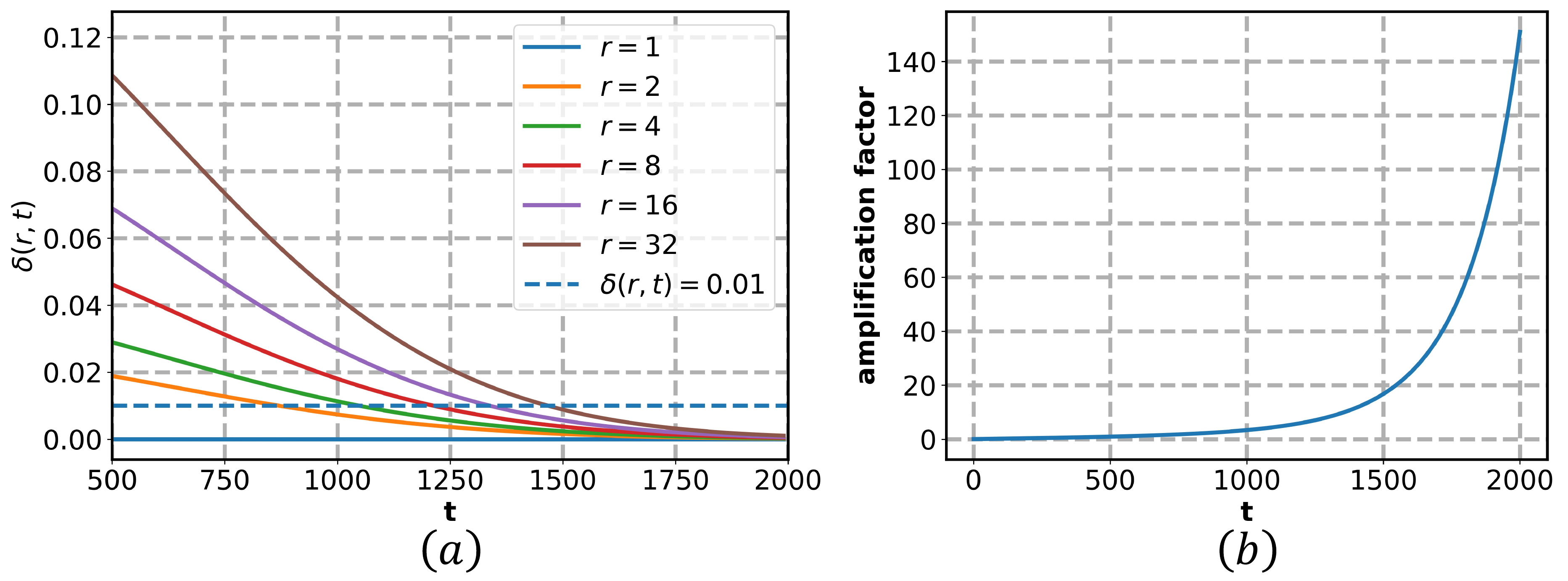}
    \end{center}
    
   \caption{(a) $\delta(r, t):=\vert(\sqrt{\bar{\alpha}_{t}}\mathbf{x}_{0}+\sqrt{1-\bar{\alpha}_{t}}\epsilon)-(\sqrt{\bar{\alpha}_{t}}(\mathbf{x}_{0}\downarrow_{r}\uparrow_{r})+\sqrt{1-\bar{\alpha}_{t}}\epsilon)\vert$ for different$(r,t)$, in which $\downarrow_{r}(\uparrow_{r})$ means downsampling (upsampling) with a scale factor of $r$. (b) Amplification factor $\frac{\sqrt{1-\Bar{\alpha}_{t}}}{\sqrt{\bar{\alpha_{}}_{t}}}$ for different $t$. \textbf{Please zoom in for the best view.}} 
    \label{fig:amplify_scale}
  \end{figure}

\subsection{Motivation}
\label{PyDiff_moti}
\noindent \textbf{Reason for Using Diffusion Models.}
Diffusion models~\cite{ho2020denoising,song2020score} have recently shown their talents in image generation, which can generate more realistic details through a sequence of refinements. Therefore, we introduce diffusion models to low-light image enhancement for better restoring noise-covered details.

\noindent \textbf{Shortcomings of Diffusion Models.}
Diffusion models have often been criticized for their slow inference speed. Many methods~\cite{song2020denoising,nichol2021improved,ho2022cascaded,saharia2022image} have been proposed to accelerate the reverse process of diffusion models. However, they still suffer from one drawback: the resolution or feature dimension is invariant in \textbf{one} reverse process of diffusion models, which limits the speed.

\noindent \textbf{Constant Resolution is Not Necessary.}
Taking resolution as an example, we note that constant resolution is not necessary for \textbf{one} reverse process. Fig.~\ref{fig:moti}(a) indicates that the first half of the reverse process can be performed at a lower resolution, which does not affect the details generated at the end. 

\noindent \textbf{The Effect of Noisy Sampling.}
Furthermore, Fig.~\ref{fig:moti}(b) shows that if global degradation (\eg, RGB shift) occurs in the first half (\ie, sampling result has more noise) of the reverse process, the second half (\ie, sampling result has less noise) will not be able to correct it. Fig.~\ref{fig:moti} demonstrates that noisy sampling (\eg, sampling in the first half of inverse processes) in diffusion models usually does not affect the final details, mainly recovering global information such as brightness and hue. \emph{Therefore, PyDiff can perform noisier sampling at a lower resolution while ensuring the global information can be recovered correctly.}

\subsection{Pyramid Diffusion}
As shown in Fig.~\ref{fig:overview}, PyDiff uses a novel pyramid diffusion method to iterate in a pyramid resolution style. Performing nosier sampling at a lower resolution can make the reverse process faster and provide the network with a larger receptive field, which is beneficial for recovering global information. In this section, we introduce the proposed pyramid diffusion.

\noindent \textbf{Downsampling Schedule.}
Similar to the noise schedule $\{\alpha\}_{t=0}^{T}$ in diffusion models, pyramid diffusion defines a downsampling schedule $\{s\}_{t=0}^{T}$, which means that the ith sampling will be performed at the resolution downsampled with a scale factor $s_{i}$. While $a_{t}\ge{a_{t+1}}$ to get bigger and bigger noise, $s_{t} \le s_{t+1}$ to get lower and lower resolution. 

\noindent \textbf{Forward Process.}
For the forward process, pyramid diffusion updates the Eq.~\eqref{eq:ddpm_forward_ori} in diffusion models as follows:
\begin{equation}
    q\left(\mathbf{x}_t | \mathbf{x}_{t-1}\right) \coloneqq \mathcal{N}\left(\mathbf{x}_t ;   \sqrt{\alpha_t} \left( \mathbf{x}_{t-1} \downarrow_{s_{t} / s_{t-1}}\right),\left(1-\alpha_t\right) \mathbf{I}\right),
\label{eq:pyr_forward_ori}
\end{equation}
where $\downarrow_{r}$ means downsampling with a scale factor of $r$. The marginal distribution $q(\mathbf{x}_{t}|\mathbf{x}_{0})$ can be expressed as follows:
\begin{equation}
q\left(\mathbf{x}_t | \mathbf{x}_{0}\right) := \mathcal{N}\left(\mathbf{x}_t ;  \sqrt{\bar{\alpha}_t} \left( \mathbf{x}_{0} \downarrow_{s_{t}} \right),\left(1-\bar{\alpha}_t\right) \mathbf{I}\right)
\label{eq:pyr_forward}
\end{equation}

\noindent \textbf{Reverse Process.}
In the case of $s_{t-1}=s_{t}$, we can derive $\mathbf{x}_{t-1}$ from $\mathbf{x}_{t}$ according to Eq.~\eqref{houyan_theta}. However, this is no longer applicable in the case of $s_{t-1}<s_{t}$ since differences in resolution. 
As shown in Fig.~\ref{fig:amplify_scale}(a), $(\mathbf{x}_{0}\downarrow_{r})\uparrow_{r}$ can serve as $x_{0}$ at noisy sampling (\ie, Larger $r$ matches noisier sampling), where $\uparrow_{r}$ means upsampling with a scale factor of $r$. 
Therefore, with well-scheduled $\{\alpha\}_{t=0}^{T}$ and $\{s\}_{t=0}^{T}$, we can take $\mathbf{y}_{\theta}(\mathbf{x}_{t})\uparrow_{s_{t}/s_{t-1}}$ as $\mathbf{x}_{0}\downarrow_{s_{t-1}}$. According to Eq.~\eqref{eq:pyr_forward}, we can further add noise to $\mathbf{x}_{0}\downarrow_{s_{t-1}}$ for deriving $\mathbf{x}_{t-1}$. Adding noise through such a method leads to inconsistency between $x_{t}$ and $x_{t-1}$. However, this inconsistency has little impact on noisy sampling, which is primarily concerned with recovering global information. To summarize, the posterior distribution of pyramid diffusion can be expressed as follows:

\begin{equation}
p_{\theta}(\mathbf{x}_{t-1}|\mathbf{x}_{t}) =
\begin{cases}

\mathcal{N}(
\mathbf{x}_{t-1} ; 
\frac{\sqrt{\bar{\alpha}_{t-1}} \beta_t}{1-\bar{\alpha}_t} \mathbf{y}_{\theta}(\mathbf{x}_{t})+\frac{\sqrt{\alpha_t}\left(1-\bar{\alpha}_{t-1}\right)}{1-\bar{\alpha}_t} \mathbf{x}_t 
\\
\quad  ,\frac{1-\bar{\alpha}_{t-1}}{1-\bar{\alpha}_t} \beta_{t} \mathbf{I}),  \text{ \ \quad \qquad \qquad if $s_{t}=s_{t-1}$} \\

\\
\mathcal{N}(
\mathbf{x}_{t-1} ; 
\sqrt{\Bar{\alpha}_{t-1}} (\mathbf{y}_{\theta}(\mathbf{x}_{t})\uparrow_{s_{t}/s_{t-1}})

\\
\ \ \ 
,(1-\Bar{\alpha}_{t-1}) \mathbf{I}), 
\text {  \quad \qquad \qquad if $s_{t}>s_{t-1}$}
\end{cases}
\end{equation}

\noindent \textbf{Position Encoding.}
Pyramid diffusion requires one network to process images of multiple resolutions. As the main operator of the denoising network~\cite{ho2020denoising}, convolution kernels cannot perceive the change of resolution. We consider using position encoding to guide the network. For an image $\mathbf{I}$ with a resolution of $H\times{W}$, its coordinates are $\mathbf{X}, \mathbf{Y} \in{\mathbb{R}^{H\times{W}} }$, where $\mathbf{X}_{i,j}=i$, and $\mathbf{Y}_{i,j}=j$. After normalizing $\mathbf{X}$ and $\mathbf{Y}$, we apply sinusoidal positional encoding of them to guide the network. Specifically, the position encoding is expressed as:
\begin{equation}
    pos(\mathbf{I})=[sin(\mathbf{X}), cos(\mathbf{X}), sin(\mathbf{Y}), cos(\mathbf{Y})]
\end{equation}
Convolution kernels have a constant receptive field. When dealing with images downsampled with a scale factor of $r$, the range of position encoding perceived by convolution kernels will be correspondingly expanded by $r$ times, which may tell convolution kernels the change of resolution.

\begin{figure}[tb]
    \begin{center}
        \includegraphics[width=0.45\textwidth]{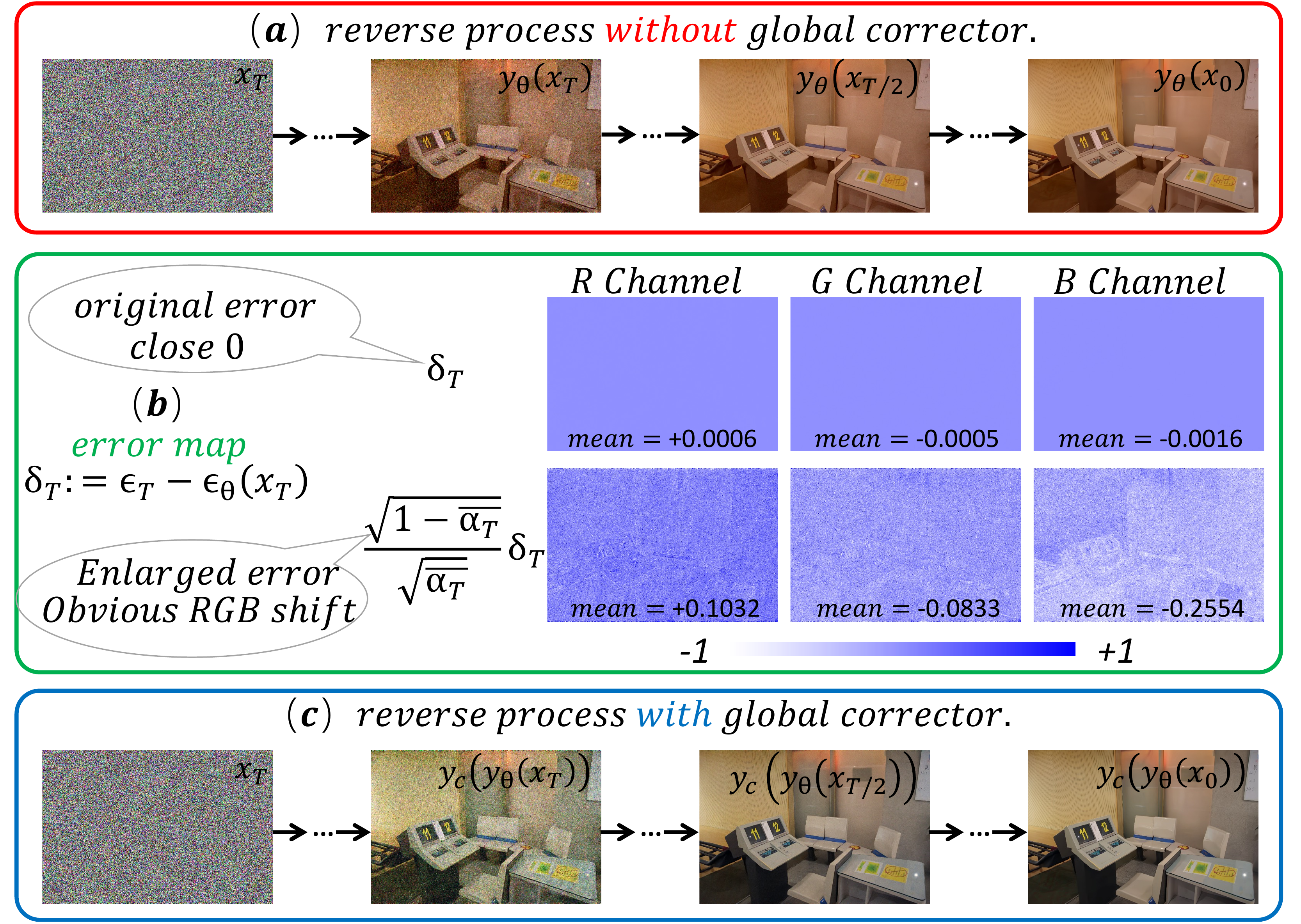}
    \end{center}
    
   \caption{$\boldsymbol{\epsilon}_{\theta}(\mathbf{x}_{t})$ is the predicted noise derived from the denoising network, and $\mathbf{y}_{\theta}(\mathbf{x}_{t})$  is an approximation of $\mathbf{x}_{0}$ calculated based on $\boldsymbol{\epsilon}_{\theta}(\mathbf{x}_{t})$. (a) Diffusion models result in significant global degradation, which appears in $\mathbf{y}_{\theta}(\mathbf{x}_{T})$ for the first time and affects subsequent sampling. (b) The original error $\boldsymbol{\delta}_{T}$ is nearly 0, but the amplification factor $\frac{\sqrt{1-\Bar{\alpha}_{T}}}{\sqrt{\bar{\alpha_{}}_{T}}}$ enlarges the error, which leads to an obvious RGB shift. (c) With the help of the global corrector, diffusion models give promising results. $\mathbf{y}_{c}(\mathbf{x})$ means using the global corrector to alleviate the global degradation in $\mathbf{x}$.} 
    \label{fig:whyHue}
  \end{figure}

\subsection{Global Corrector}
As shown in Fig.~\ref{fig:overview}, PyDiff uses the global corrector to alleviate global degradation in reverse processes. The global corrector can significantly improve performance with little additional computational consumption. In this section, we introduce the proposed global corrector.

\noindent \textbf{Global Degradation.}
When applying diffusion models to low-light image enhancement, we found them sometimes result in significant global degradation, as shown in Fig.~\ref{fig:whyHue}(a). This global degradation looks like a shift in the RGB channels, similar to the RGB shift shown in Fig.~\ref{fig:moti}(b).

\noindent \textbf{Cause of Global Degradation.}
\label{sec:cause}
Looking back on Eq.~\eqref{eq:cal_x0}, we can rewrite it as follows:
\begin{linenomath}
\begin{align}
    \mathbf{y}_{\theta}(\mathbf{x}_{t})&:=\frac{1}{\sqrt{\bar{\alpha_{}}_{t}}}(\mathbf{x}_{t}-\sqrt{1-\bar{\alpha}_{t}} (\boldsymbol{\epsilon}_{t}-\boldsymbol{\delta}_{t}) )\\
    &:=\mathbf{x}_{0} + \frac{\sqrt{1-\Bar{\alpha}_{t}}}{\sqrt{\bar{\alpha_{}}_{t}}}\boldsymbol{\delta}_{t},
\end{align}
\end{linenomath}
where $\boldsymbol{\epsilon}_{t}$ is the actual noise in $\mathbf{x}_{t}$, and $\boldsymbol{\delta}_{t}$ is the error between $\boldsymbol{\epsilon}_{t}$ and $\boldsymbol{\epsilon}_{\theta}(\mathbf{x}_{t})$. We found that there is a coefficient $\frac{\sqrt{1-\Bar{\alpha}_{t}}}{\sqrt{\bar{\alpha_{}}_{t}}}$ in front of the error $\boldsymbol{\delta}_{t}$. When $t$ is relatively large, this coefficient will also be large, as shown in Fig.~\ref{fig:amplify_scale}(b). As demonstrated in Fig.~\ref{fig:whyHue}(b), the original error $\boldsymbol{\delta}_{T}$ is small, but the coefficient $\frac{\sqrt{1-\Bar{\alpha}_{T}}}{\sqrt{\bar{\alpha_{}}_{T}}}$ enlarges the error and leads to an obvious RGB shift (\eg, a significant gain in the R channel). As shown in Fig.~\ref{fig:whyHue}(a), the denoising network treats the image under global degradation as usual and only performs its denoising duties, which can not eliminate the global degradation.

\noindent \textbf{Design of Global Corrector.}
We add a global corrector to alleviate the global degradation that denoising networks can not notice. The design of the global corrector needs to meet the following requirements: \textbf{1}) The global corrector should alleviate global degradation while preserving generated edges and textures. \textbf{2}) The global corrector is lightweight and fast. Inspired by CSRNet~\cite{he2020conditional}, we design an efficient global corrector that performs pixel-independent retouching based on global conditions. Please refer to the supplementary materials for the specific design of the global corrector. To sample with a global corrector, we update Eq.~\eqref{houyan_theta} as follows: 
\begin{equation}
    p_{\theta,c}\left(\mathbf{x}_{t-1} | \mathbf{x}_t\right):=\mathcal{N}\left(\mathbf{x}_{t-1} ; \tilde{\boldsymbol{\mu}}_t\left(\mathbf{x}_t, \mathbf{y}_{c}(\mathbf{y}_{\theta}(\mathbf{x}_{t}))\right), \tilde{\beta}_t \mathbf{I}\right)
\label{houyan_theta_c}
\end{equation}
where $\mathbf{y}_{c}(\mathbf{x})$ means using the global corrector to alleviate the global degradation in $\mathbf{x}$. Fig.~\ref{fig:whyHue}(c) shows that the global corrector can alleviate global degradation while maintaining generated edges and textures. 

\noindent \textbf{Correction Threshold.}
As shown in Fig.~\ref{fig:amplify_scale}(b), the amplification factor $\frac{\sqrt{1-\Bar{\alpha}_{t}}}{\sqrt{\bar{\alpha_{}}_{t}}}$ will gradually decrease to 0 as t decreases. When the amplification factor is small enough, it will no longer amplify the error of the denoising network. Therefore, We set a correction threshold $\gamma$, and the global corrector is needed only when the $\frac{\sqrt{1-\Bar{\alpha}_{t}}}{\sqrt{\bar{\alpha_{}}_{t}}} > \gamma$. We set $\gamma=1$.

\begin{algorithm}[tb]
    \caption{Training}
    \label{alg:train}
    \begin{algorithmic}[1] 
        \STATE \textbf{Input}: noise schedule $\alpha$, downsampling schedule $s$, correction threshold $\gamma$, denoising network $\theta$, global corrector $c$, low/normal-light image pairs $q(\mathbf{x}_{low},\mathbf{y})$.
        \REPEAT
        \STATE  Sample $(\mathbf{x}_{low}, \mathbf{y}) \sim{q(\mathbf{x}_{low}, \mathbf{y})}$.
        \STATE  Sample $t \sim{Uniform({1,...,T})}$
        \STATE  Sample $\boldsymbol{\epsilon} \sim{\mathcal{N}(0, I)}$
        \STATE $\mathbf{x}_{t}=  \sqrt{\bar{\alpha}_t} (\mathbf{y}\downarrow_{s_{t}}) +\sqrt{1-\bar{\alpha}_t}\boldsymbol{\epsilon}$ \\$\triangleright$ $\downarrow_{r}$ means downsampling with a scale factor of $r$.
        
        \STATE Take gradient descent step on \\
        $\quad \quad \nabla_\theta\left\|\boldsymbol{\epsilon}-\boldsymbol{\epsilon}_\theta\left(\mathbf{x}_{t},(\mathbf{x}_{low}\downarrow_{s_{t}})\right)\right\|_1$
        \IF{$\frac{\sqrt{1-\bar{\alpha}_{t}}}{\sqrt{\bar{\alpha}_{t}}} > \gamma$}
        \STATE Take gradient descent step on \\
                $\quad  \quad \nabla_c \left\|{(\mathbf{y}\downarrow_{s_{t}})}-{\mathbf{y}}_{c}\left(\mathbf{y}_{\theta}(\mathbf{x}_{t}, (\mathbf{x}_{low}\downarrow_{s_{t}}))\right)\right\|_1$
        \ENDIF
        \UNTIL{converged}
    \end{algorithmic}
\end{algorithm}

\begin{algorithm}[tb]
    \caption{Sampling}
    \label{alg:test}
    \begin{algorithmic}[1] 
    \STATE \textbf{Input}: noise schedule $\alpha$, downsampling schedule $s$, correction threshold $\gamma$, denoising network $\theta$, global corrector $c$, low-light image $\mathbf{x}_{low}$.
        \STATE Sample $\mathbf{x}_T \sim \mathcal{N}(\mathbf{0}, \mathbf{I})$
        \FOR{$t = T,...,1$}
        \STATE $\mathbf{y}'= \mathbf{y}_{\theta}(\mathbf{x}_{t}, \mathbf{x}_{low})$
        \STATE $\mathbf{y}'= \mathbf{y}_{c}(\mathbf{y}') \text { if } \frac{\sqrt{1-\bar{\alpha}_{t}}}{\sqrt{\bar{\alpha}_{t}}} > \gamma, \text{ else } \mathbf{y}'=\mathbf{y}'$
        \STATE Sample $\boldsymbol{\epsilon} \sim \mathcal{N}(\mathbf{0}, \mathbf{I}) \text { if } t>1, \text { else } \boldsymbol{\epsilon}=\mathbf{0}$
        \IF{$s_{t} > s_{t-1}$}
        \STATE $\mathbf{x}_{t-1} =\sqrt{\bar{\alpha}_{t-1}}(\mathbf{y}'\uparrow_{s_{t}/s_{t-1}})+\sqrt{1-\bar{\alpha}_{t-1}}\boldsymbol{\epsilon}$
        \\$\triangleright$ $\uparrow_{r}$ means upsampling with a scale factor of $r$.
        \ELSE
        
        \STATE $\mathbf{x}_{t-1}= \frac{\sqrt{\bar{\alpha}_{t-1}} \beta_t}{1-\bar{\alpha}_t} \mathbf{y}' +\frac{\sqrt{\alpha_t}\left(1-\bar{\alpha}_{t-1}\right)}{1-\bar{\alpha}_t} \mathbf{x}_t+\sqrt{\frac{1-\bar{\alpha}_{t-1}}{1-\bar{\alpha}_t} \beta_{t}}\boldsymbol{\epsilon}$
        \ENDIF
        \ENDFOR
        \RETURN $x_{0}$
    \end{algorithmic}
\end{algorithm}
\subsection{Training and Sampling}
To better demonstrate the overall framework of our PyDiff, we omit some details (\eg, position encoding) when describing the algorithm.

\noindent \textbf{Training.}
Algorithm~\ref{alg:train} shows the specific training procedure of the proposed PyDiff. The global corrector aims to alleviate the global degradation that the denoising network cannot notice, and it will not impact the denoising network.

\noindent \textbf{Sampling.}
Algorithm~\ref{alg:test} shows the specific sampling procedure of the proposed PyDiff. PyDiff can be easily combined with DDIM~\cite{song2020denoising} or DDPM+~\cite{nichol2021improved} to achieve further speedup.

\noindent \textbf{Training Loss.}
As described in algorithm~\ref{alg:train}, we use the simple L1 loss to optimize the denoising network and global corrector without additional optimization objectives.

\section{Experiments}
\begin{figure*}[tb]
    \centering
    \setlength{\tabcolsep}{0pt} 
    \renewcommand{\arraystretch}{0.0} 
    \begin{tabular}{cccccccccc}
{\small Input} & {\small IAT} & {\small KIND}  & {\small KIND++} & {\small BREAD}  & {\small NE} & {\small HWMNet}  & {\small LLFLOW} & {\small PyDiff} & {\small Reference}\\
{\includegraphics[width=0.1\linewidth, height=0.1\linewidth]{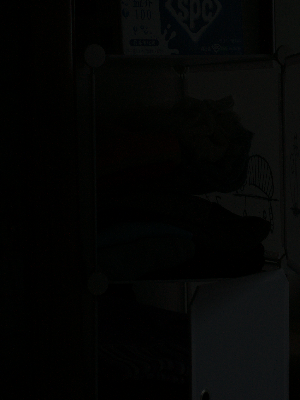}} &
{\includegraphics[width=0.1\linewidth, height=0.1\linewidth]{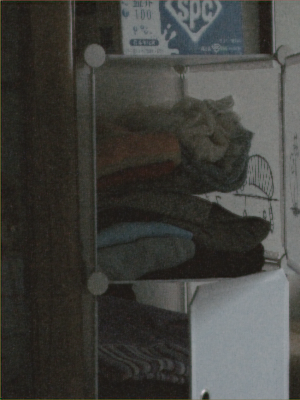}} &
{\includegraphics[width=0.1\linewidth, height=0.1\linewidth]{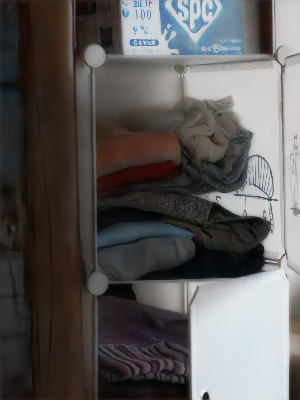}} &
{\includegraphics[width=0.1\linewidth, height=0.1\linewidth]{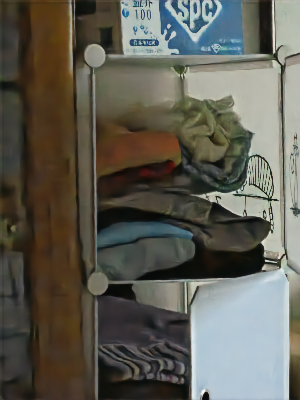}}&
{\includegraphics[width=0.1\linewidth, height=0.1\linewidth]{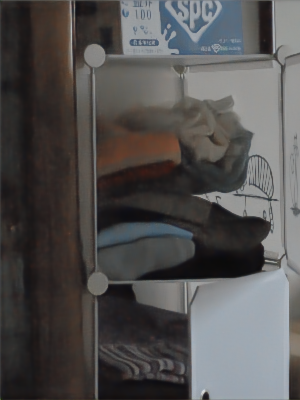}}&
{\includegraphics[width=0.1\linewidth, height=0.1\linewidth]{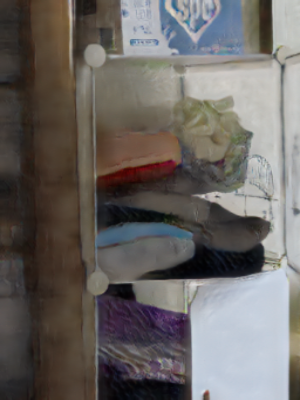}}&
{\includegraphics[width=0.1\linewidth, height=0.1\linewidth]{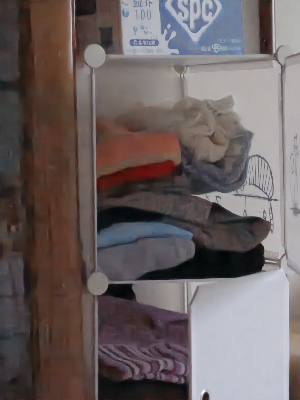}}&
{\includegraphics[width=0.1\linewidth, height=0.1\linewidth]{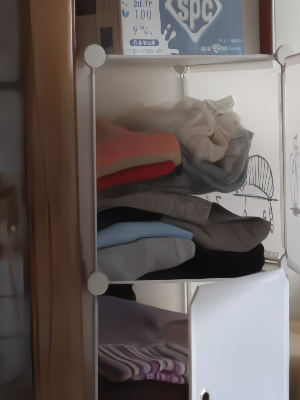}}&
{\includegraphics[width=0.1\linewidth, height=0.1\linewidth]{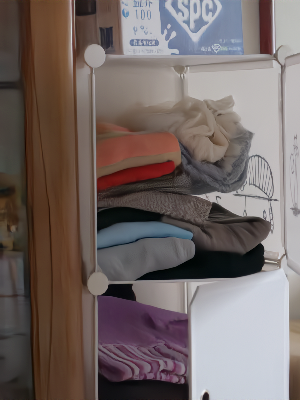}}&
{\includegraphics[width=0.1\linewidth, height=0.1\linewidth]{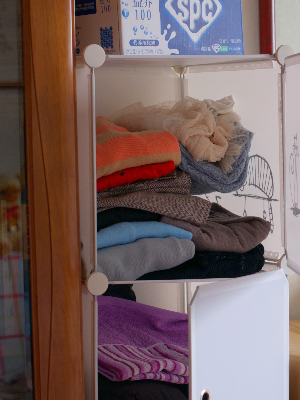}}
\\
\end{tabular}
    \caption{Qualitative comparison with state-of-the-art methods on the LOL dataset. It can be seen that IAT~\protect\cite{cui2022illumination} cannot even restore the correct brightness, and the results generated by KIND++~\protect\cite{zhang2021beyond}, BREAD~\protect\cite{guo2022low}, NE~\protect\cite{jin2022unsupervised}, and HWMNet~\protect\cite{fan2022half} have apparent artifacts, while the KIND~\protect\cite{zhang2019kindling} cannot restore colors well. LLFLOW~\protect\cite{wang2022low} gives a not-bad result, but its result can be too smooth, and the colors of some items need to be more accurately restored. PyDiff exhibits the best result, which restores the correct color and preserves the details covered by noise. \textbf{Please zoom in for the best view.}}
    \label{fig:lol-result}
\end{figure*}

\begin{figure*}[tb]
    \centering
    \begingroup
    \setlength{\tabcolsep}{0pt} 
    \renewcommand{\arraystretch}{0.} 
    \begin{tabular}{cccccccccc}
{\small Input} & {\small IAT} & {\small KIND}  & {\small KIND++} & {\small BREAD}  & {\small NE} & {\small HWMNet}  & {\small LLFLOW} & {\small PyDiff} & {\small Reference}\\
{\includegraphics[width=0.1\linewidth, height=0.1\linewidth]{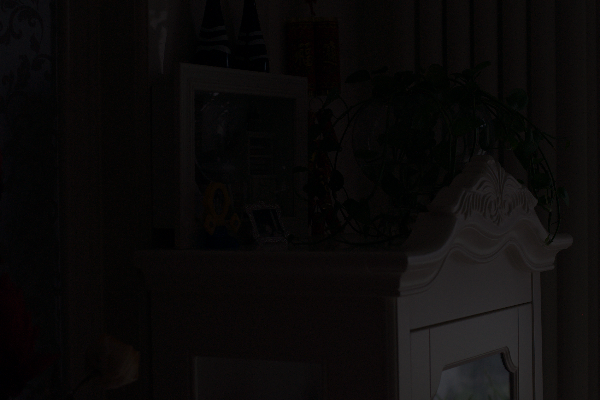}} &
{\includegraphics[width=0.1\linewidth, height=0.1\linewidth]{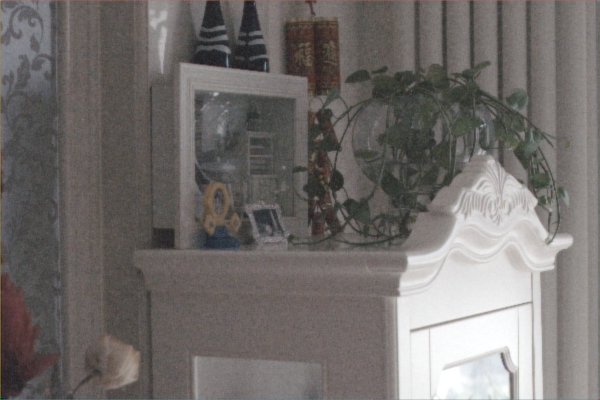}} &
{\includegraphics[width=0.1\linewidth, height=0.1\linewidth]{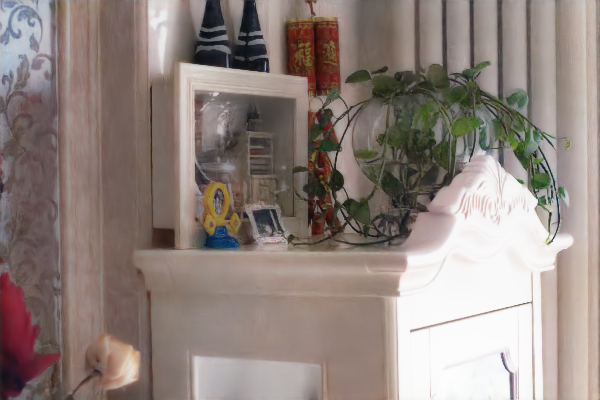}} &
{\includegraphics[width=0.1\linewidth, height=0.1\linewidth]{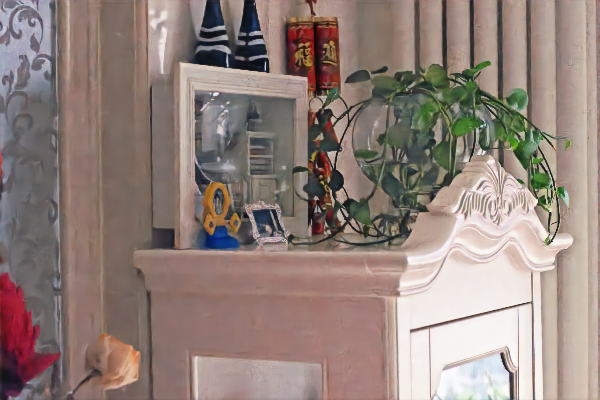}}&
{\includegraphics[width=0.1\linewidth, height=0.1\linewidth]{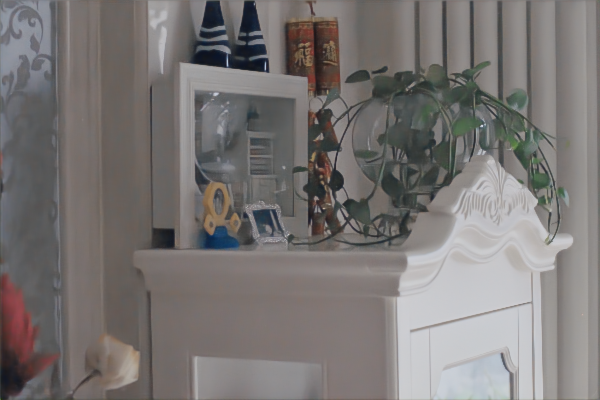}}&
{\includegraphics[width=0.1\linewidth, height=0.1\linewidth]{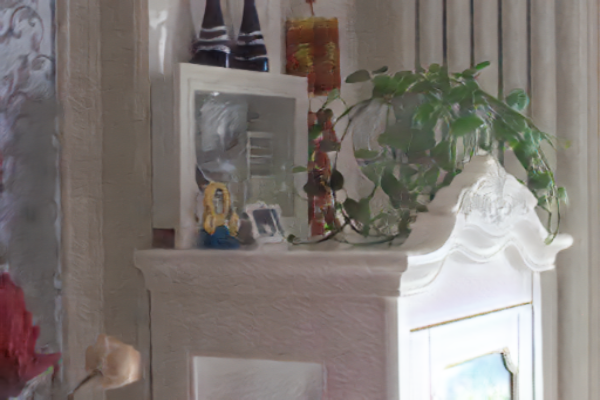}}&
{\includegraphics[width=0.1\linewidth, height=0.1\linewidth]{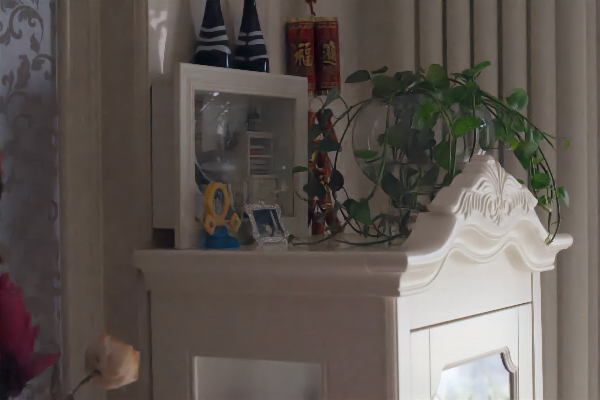}}&
{\includegraphics[width=0.1\linewidth, height=0.1\linewidth]{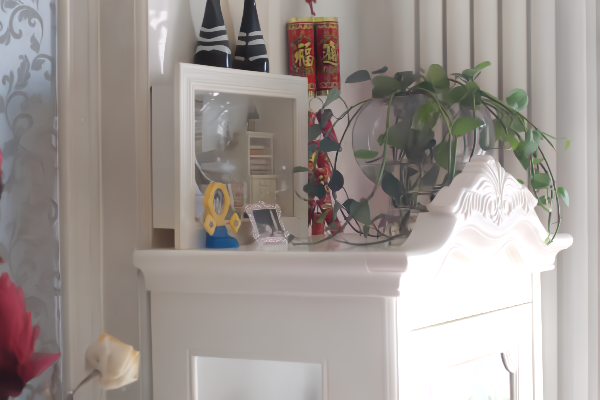}}&
{\includegraphics[width=0.1\linewidth, height=0.1\linewidth]{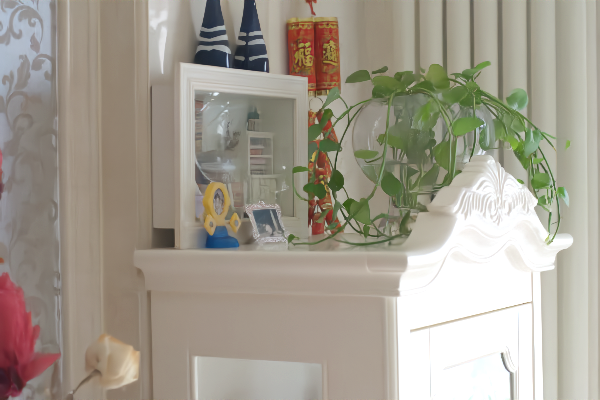}}&
{\includegraphics[width=0.1\linewidth, height=0.1\linewidth]{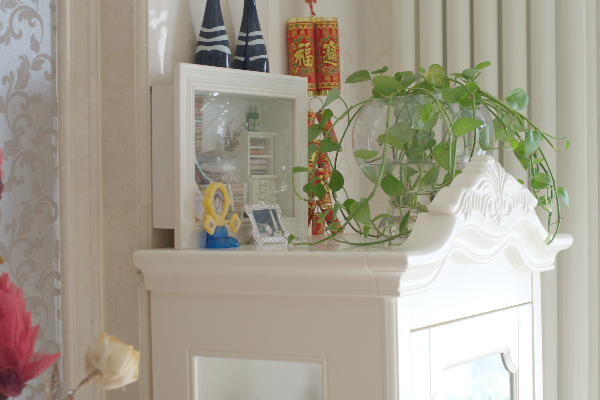}}
\\

{\includegraphics[width=0.1\linewidth, height=0.1\linewidth]{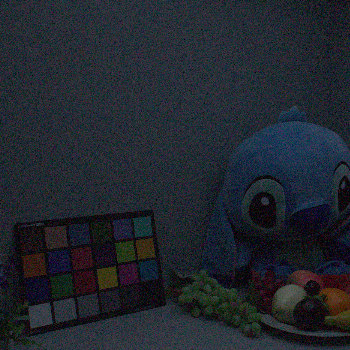}} &
{\includegraphics[width=0.1\linewidth, height=0.1\linewidth]{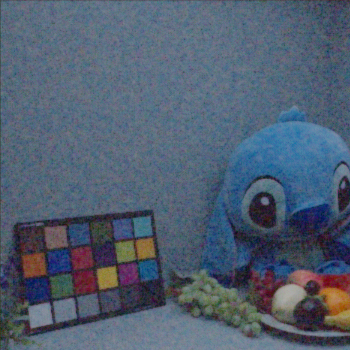}} &
{\includegraphics[width=0.1\linewidth, height=0.1\linewidth]{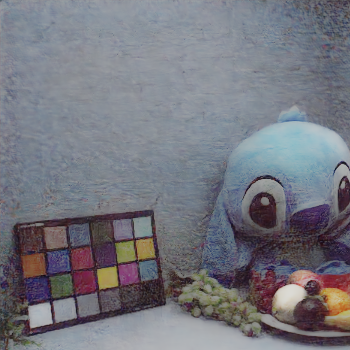}} &
{\includegraphics[width=0.1\linewidth, height=0.1\linewidth]{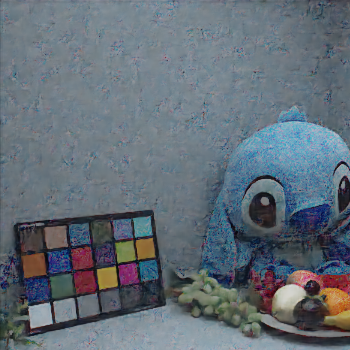}}&
{\includegraphics[width=0.1\linewidth, height=0.1\linewidth]{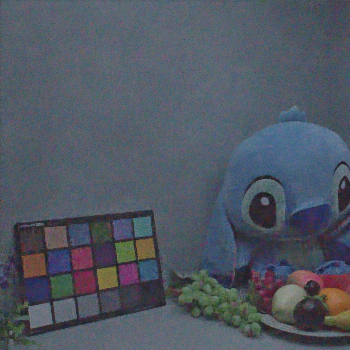}}&
{\includegraphics[width=0.1\linewidth, height=0.1\linewidth]{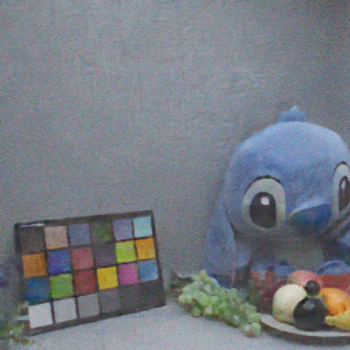}}&
{\includegraphics[width=0.1\linewidth, height=0.1\linewidth]{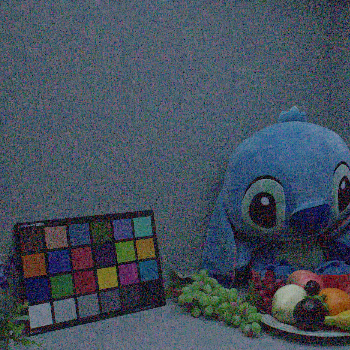}}&
{\includegraphics[width=0.1\linewidth, height=0.1\linewidth]{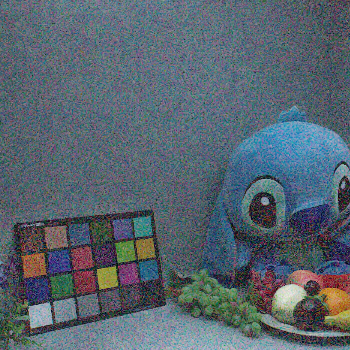}}&
{\includegraphics[width=0.1\linewidth, height=0.1\linewidth]{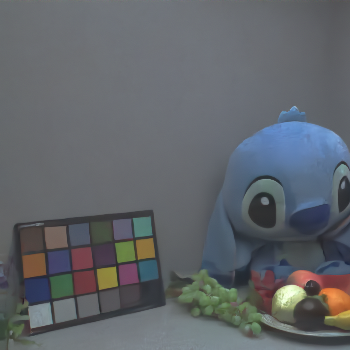}}
&{\includegraphics[width=0.1\linewidth, height=0.1\linewidth]{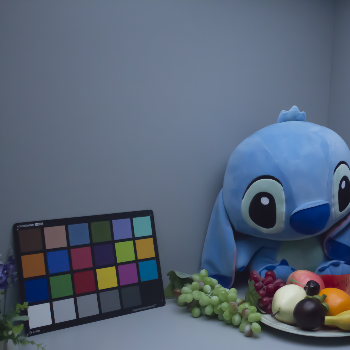}}
\\

\end{tabular}
\endgroup
    \caption{Qualitative comparison with state-of-the-art methods on the REAL PART of the LOLV2 dataset. \textbf{Please zoom in for the best view.}}%
    \label{fig:lolv2-real-result}
\end{figure*}

\begin{table}[tb]  

        \setlength{\tabcolsep}{3pt}
        \renewcommand{\arraystretch}{0.9} 
	\centering
	\small
		\scalebox{1.0}{
			\begin{tabular}{lccc}
				\toprule
				Methods  & PSNR $\bm{\uparrow}$ & SSIM  $\bm{\uparrow}$ & LPIPS    $\bm{\downarrow}$ 
				\\ \midrule
				Zero-DCE~\cite{guo2020zero} &$14.86$ & $0.54$ & $0.33$ \\
				  EnlightenGAN~\cite{jiang2021enlightengan} &$17.48$ & $0.65$ & $0.32$ \\
				KinD~\cite{zhang2019kindling} &$20.87$ & $0.80$ & $0.17$ \\
                    KinD++~\cite{zhang2021beyond} &$21.30$ & $0.82$ & $0.16$ \\
                    RCTNet~\cite{kim2021representative} &$22.67$ & $0.79$ & $-$ \\
                    Bread~\cite{guo2022low} &$22.96$ & $0.84$ & $0.16$ \\
                    NE~\cite{jin2022unsupervised} &$21.52$ & $0.76	$ & $0.24$ \\
                    IAT~\cite{cui2022illumination} &$23.38$ & $0.81$ & $0.26$ \\
                    HWMNet~\cite{fan2022half} &$24.24$ & $0.85$ & $0.12$ \\
                    LLFLOW~\cite{wang2022low} &$24.99$ & $0.92$ & $0.11$ \\
				\textbf{PyDiff (ours)} & \textbf{27.09} & \textbf{0.93} & \textbf{0.10}  \\
				\bottomrule
    
		\end{tabular}
                    } 
            
         \caption{Quantitative results on the LOL dataset in terms of PSNR, SSIM, and LPIPS. $\uparrow$ ($\downarrow$) denotes that larger (smaller) values lead to better quality.}
	\label{tab:LOL}

\end{table}

\subsection{Setup}

\noindent \textbf{Dataset.}
We conduct experiments on LOL~\cite{wei2018deep} and LOLV2~\cite{yang2021sparse} datasets. The LOLV2 dataset contains two parts, REAL and SYNC. REAL PART has noise distributions not present in the LOL dataset, and SYNC PART has illumination distributions not present in the LOL dataset. For supervised learning methods involved in comparison, we use their pre-trained model only trained on the LOL dataset. Correspondingly, we train PyDiff only on the LOL dataset too. For unsupervised learning methods, we use their released pre-trained models no matter how they train.

\noindent \textbf{Schedules.}
First of all, we set $T=2000$. For the noise schedule, we decrease $\alpha_{t}$ linearly from $\alpha_{1}=0.999999$ to $\alpha_{T}=0.99$. For the downsampling schedule, our default setting is to set $\{s\}_{t=1}^{T/2}=1$ and $\{s\}_{t=T/2}^{T}=2$, and we will experiment with more schedules in ablation studies.

\noindent \textbf{Training.}
We set the patch size to $192 \times 288$ and the batch size to $16$. We use the Adam optimizer with an initial learning rate of $1\times{10^{-4}}$ for 320k iterations and halve the learning rate at 50k, 75k, 100k, 150k, and 200k. The optimizer does not use weight decay. We complete training on two NVIDIA GeForce RTX 3090s.

\noindent \textbf{Evaluation.}
Combined with DDIM~\cite{song2020denoising}, PyDiff requires only \textbf{4} iterations to obtain results comparable to other SOTA methods.

\noindent \textbf{Other Details.}
The reverse process is conditional on low-light images $x_{low}$, low-light images after histogram equalization $hiseq(x_{low})$, and position encoding $pos(x_{low})$. We use the method of concatenating to achieve conditional sampling~\cite{saharia2022image,saharia2022palette}. During training, we swap the concatenating order of $x_{low}$ and $hiseq(x_{low})$ with a probability of 0.5. \textbf{Please refer to the supplementary materials for model configuration and details of the global corrector.}

\begin{table}[tb]  

        \setlength{\tabcolsep}{3pt}
        \renewcommand{\arraystretch}{0.9} 
	\centering
	\small
		\scalebox{1.0}{
			\begin{tabular}{lccc}
				\toprule
				Methods  & PSNR $\bm{\uparrow}$ & SSIM  $\bm{\uparrow}$ & LPIPS    $\bm{\downarrow}$ 
				\\ \midrule

                Zero-DCE~\cite{guo2020zero} &$13.65$ & $0.246$ & $0.98$ \\
				KinD~\cite{zhang2019kindling} &$20.40$ & $0.652$ & $0.50$ \\
                    KinD++~\cite{zhang2021beyond} &$20.15$ & $0.678$ & $0.47$ \\
                    NE~\cite{jin2022unsupervised} &$21.12$ & $0.767$ & $0.46$ \\
                    IAT~\cite{cui2022illumination} &$21.43$ & $0.638$ & $0.60$ \\
                    Bread~\cite{guo2022low} &$22.54$ & $0.762$ & $0.44$ \\
                    HWMNet~\cite{fan2022half} &$22.40$ & $0.622$ & $0.56$ \\
                    LLFLOW~\cite{wang2022low} &$21.60$ & $0.643$ & $0.53$ \\
				\textbf{PyDiff (ours)} & \textbf{24.01} & \textbf{0.876} & \textbf{0.23}  \\
				\bottomrule
		\end{tabular}
                    }
                    
	\caption{Quantitative results on the LOLV2 REAL PART in terms of PSNR, SSIM, and LPIPS. All methods involved in the comparison were not retrained on the corresponding training set. $\uparrow$ ($\downarrow$) denotes that larger (smaller) values lead to better quality.}
	\label{tab:LOLV2_real}
\end{table}

\begin{table}[tb]  

        \setlength{\tabcolsep}{3pt}
        \renewcommand{\arraystretch}{0.9} 
	\centering
	\small
		\scalebox{1.0}{
			\begin{tabular}{lccc}
				\toprule
				Methods  & PSNR $\bm{\uparrow}$ & SSIM  $\bm{\uparrow}$ & LPIPS    $\bm{\downarrow}$ 
				\\ \midrule
                    KIND~\cite{zhang2019kindling} &$18.32$ & $0.822$ & $0.25$ \\KIND++~\cite{zhang2021beyond} &$19.44$ & $0.830$ & $0.23$ \\IAT~\cite{cui2022illumination} &$19.18$ & $0.813$ & $0.29$ \\
                    Bread~\cite{guo2022low} &$19.28$ & $0.831$ & $0.24$ \\
                    HWMNet~\cite{fan2022half} &$18.79$ & $0.817$ & $0.24$ \\
                    LLFLOW~\cite{wang2022low} &$19.15$ & $0.860$ & $\textbf{0.22}$ \\
				\textbf{PyDiff (ours)} & \textbf{19.60} & \textbf{0.878} & \textbf{0.22}  \\
				\bottomrule
		\end{tabular}
                    } 
                    
	\caption{Quantitative results on the LOLV2 SYNC PART in terms of PSNR, SSIM, and LPIPS. All methods involved in the comparison were not retrained on the corresponding training set. $\uparrow$ ($\downarrow$) denotes that larger (smaller) values lead to better quality.}
	\label{tab:LOLV2_sync}
\end{table}

\subsection{Comparsion with SOTA Methods}
\noindent \textbf{LOL Dataset.}
We first compare PyDiff with SOTA methods on the LOL dataset. The quantitative results are shown in Tab.~\ref{tab:LOL}. PyDiff outperforms other methods in all three metrics: PSNR, SSIM~\cite{wang2004image}, and LPIPS~\cite{zhang2018unreasonable}. Beating second place by \textbf{2.1} points on PSNR shows that PyDiff can recover more accurate colors. Surpassing second place by \textbf{1} point on SSIM shows that PyDiff accurately preserves more high-frequency details. Exceeding second place by \textbf{1} point on LPIPS shows that PyDiff gives more eye-pleasing results. Fig.~\ref{fig:lol-result} shows qualitative comparisons with other methods, where PyDiff exhibits the best result.

\begin{table}[tb]
\setlength{\tabcolsep}{3pt}
\renewcommand{\arraystretch}{0.9} 
\small
\centering

\begin{tabular}{lcccccc}
\toprule

\multirow{2}{*}{}
& \multicolumn{2}{c}{Hyperparameters}  & \multicolumn{4}{c}{Metrics} \\

\cmidrule(lr){2-3} \cmidrule(lr){4-7}

 & $schedule$ & $pe$ & PSNR$\bm{\uparrow}$ & SSIM$\bm{\uparrow}$
& LPIPS$\bm{\downarrow}$ & FPS$\bm{\uparrow}$ \\

\midrule

LLFLOW & - & -  & $20.70$  & $0.763$ & $0.36$ & $1.94$  \\

\midrule

\multirow{6}{*}{PyDiff} 
& $[1,1,1,1]$ & $YES$  & $22.11$  & $0.878$ & $0.22$ & $2.35$  \\
& $[1,1,2,2]$ & $YES$  & \textbf{22.17}  & \textbf{0.881} & \textbf{0.22} & $3.62$  \\
& $[1,1,2,2]$  & $NO$  & $21.96$  & $0.878$ & $0.22$ & $3.62$ \\
    



 & $[1,1,2,4]$ & $YES$  & $22.02$  & $0.879$ & $0.22$ & 3.81  \\
 & $[1,2,2,2]$ & $YES$  & $22.15$  & $0.877$ & $0.22$ & 4.97  \\
 & $[1,2,4,8]$ & $YES$  & $22.12$  & $0.875$ & $0.22$ & \textbf{5.86}  \\

\bottomrule

\end{tabular}
\caption{Ablation study on the pyramid diffusion. $schedule$ stands for downsampling schedule, while $pe$ means position encoding.}
\label{tab:pyr-abl}
\end{table}

\noindent \textbf{LOLV2 REAL PART.}
Since the test set of LOLV2 REAL PART overlaps with the training set of the LOL dataset, we combine the training set and test set of LOLV2 REAL PART and filter out the overlapping parts with the LOL training set by the ID of the images. For the filtered images, we sort them by ID and select 100 (\ie, the same size as the original test set) images with the smallest ID as the test set of LOLV2 REAL PART. \emph{Many of the selected images were taken at ISOs not included in the LOL training set, which is a good test of the model's ability to deal with unseen noise.} Tab.~\ref{tab:LOLV2_real} shows the quantitative comparison with other SOTA methods on LOLV2 REAL PART. As PyDiff can deal with unseen noise better, it outperforms second place by \textbf{10.9} points on SSIM and \textbf{21} points on LPIPS. As shown in the second row of Fig.~\ref{fig:lolv2-real-result}, other SOTA methods give results with significant noise, while PyDiff can remove noise well. At the same time, the first row of Fig.~\ref{fig:lolv2-real-result} also shows that PyDiff can better restore images with different exposure times.

\noindent \textbf{LOLV2 SYNC PART.}
LOLV2 SYNC PART contains many illumination distributions that the LOL dataset does not have, and the scenarios in it are entirely different from the LOL dataset, which can test models' generalization. Tab.~\ref{tab:LOLV2_sync} shows the quantitative comparison between PyDiff and other SOTA methods on LOLV2 SYNC PART. PyDiff shows competitive results and achieves first place in performance (\eg, \textbf{1.8} points higher than second place on SSIM), which demonstrates the generalization of PyDiff. Supplementary materials will show the qualitative comparison with other SOTA methods on LOLV2 SYNC PART.

\subsection{Ablation Study}
In this section, we conduct ablation studies on the main components of PyDiff to observe their impact on performance. The score for this section is calculated by combining the performance on the LOL dataset, LOLV2 REAL PART, and LOLV2 SYNC PART, which gives a better indication of the effectiveness of a component. FPS is measured on the LOL dataset (\ie, the resolution is $400\times{600}$).

\noindent \textbf{Downsampling Schedules.}
In Tab.~\ref{tab:pyr-abl}, schedule $[1,1,1,1]$ represents vanilla diffusion models, which sample at a constant resolution. Our default setting, schedule $[1,1,2,2]$, performs noisy sampling at a 1/2 resolution.
Schedule $[1,1,2,2]$ is $54\%$ faster while slightly outperforming the schedule $[1,1,1,1]$. 
Furthermore, our investigation revealed that faster schedules (\eg, $[1,1,2,4]$, $[1,2,2,2]$, and $[1,2,4,8]$) produce comparable results to the vanilla schedule $[1,1,1,1]$. These findings indicate that noisier sampling can be performed at a lower resolution, while still maintaining high performance.

\noindent \textbf{Position Encoding.}
Tab.~\ref{tab:pyr-abl} shows that the position encoding boosts PSNR and SSIM for PyDiff, which may tell networks about the change of resolution.

\begin{table}[tb]
\small
\setlength{\tabcolsep}{2pt}
\renewcommand{\arraystretch}{0.9} 
\centering
\begin{tabular}{cccccc}
\toprule

 \multicolumn{2}{c}{Hyperparameters}  & \multicolumn{4}{c}{Metrics} \\

\cmidrule(lr){1-2} \cmidrule(lr){3-6}

  $batch$ & $gc$ & PSNR$\bm{\uparrow}$ & SSIM$\bm{\uparrow}$
& LPIPS$\bm{\downarrow}$ & FPS$\bm{\uparrow}$ \\

\midrule

 $16$ & $YES$  & $22.17$  & $0.881$ & $0.22$ & $3.62$  \\
 \textbf{(default)} & $NO$  & $21.65$  & $0.875$ & $0.22$ & $3.70$ \\

\midrule

 \multirow{2}{*}{$8$} & $YES$  & $21.98$  & $0.873$ & $0.22$ & $3.62$  \\
   & $NO$  & $20.74$  & $0.861$ & $0.25$ & $3.70$ \\
    
\midrule
    
     \multirow{2}{*}{$4$}& $YES$  & $21.42$  & $0.868$ & $0.25$ & $3.62$  \\
     & $NO$  & $18.35$  & $0.843$ & $0.35$ & $3.70$ \\

\bottomrule
\end{tabular}
\caption{Ablation study on the global corrector. $batch$ stands for batch size, while $gc$ means global corrector.}
\label{tab:gc-abl}
\end{table}

\noindent \textbf{Effectiveness of Global Corrector.} 
Tab.~\ref{tab:gc-abl} shows that the global corrector can bring improvements to PyDiff under various settings. Fig.~\ref{fig:overview} and Fig.~\ref{fig:whyHue}(a)(c) show that the global corrector effectively alleviates global degradation. Furthermore, we can see from Tab.~\ref{tab:gc-abl} that the global corrector gives little additional computational consumption to PyDiff. 

\noindent \textbf{Robustness to Batch Size.}
Tab.~\ref{tab:gc-abl} shows that vanilla diffusion models without global corrector are very dependent on large batch size, which shows a significant performance drop when the batch size decreases. As our analysis in section~\ref{sec:cause}, we argue that this is caused by the amplification factor, which has rigorous requirements for denoising networks. This problem has been significantly improved by adding a global corrector. As shown in Tab.~\ref{tab:gc-abl}, the global corrector enhances the performance of diffusion models under $bs=4(8)$ and outperforms the ones without global corrector under $bs=8(16)$, which means that the global corrector can make diffusion models more robust to batch size and easier to train.

\noindent \textbf{Comparison with LLFLOW.} 
LLFLOW~\cite{wang2022low} used to be first place on the LOL dataset based on the normalizing flow~\cite{dinh2016density,kingma2018glow}. Both FLOWs and diffusion models are generative models that require multiple iterations. Therefore, it will be interesting to compare the speed of LLFLOW and PyDiff. According to Tab.~\ref{tab:pyr-abl}, PyDiff significantly enhances performance, achieving an 87$\%$ faster speed than LLFLOW.

\section{Conclusion}
This paper proposes PyDiff, a diffusion model based method for low-light image enhancement. PyDiff uses a novel pyramid diffusion method, which makes sampling faster than vanilla diffusion models without any performance degradation. Furthermore, PyDiff uses a global corrector to alleviate global degradations that cannot be noticed by the denoising network and significantly improves performance with little additional computational consumption. Experimentally, PyDiff shows superior effectiveness, efficiency, and generalization ability on popular benchmarks. We hope that PyDiff will serve as a strong baseline for low-light image enhancement and that the pyramid diffusion method will facilitate the application of diffusion models in more low-level vision tasks.

\noindent\textbf{Acknowledgement.} This work is partially supported by the Fundamental Research Funds for the Central Universities (No. 226-2022-00051).

\bibliographystyle{named}
\bibliography{ijcai23}

\end{document}